\documentclass[runningheads]{llncs}

\usepackage[mobile]{eccv}

\usepackage{eccvabbrv}

\usepackage{graphicx}
\usepackage{booktabs}

\usepackage[accsupp]{axessibility}  
\definecolor{nicegreen}{rgb}{0.1, 0.6, 0.2}

\usepackage{hyperref}

\usepackage{orcidlink}

\usepackage{graphicx}
\usepackage{amsmath}
\usepackage{amssymb}
\usepackage{booktabs}

\usepackage[ruled,vlined]{algorithm2e}
\usepackage{cuted}
\usepackage{fvextra}   

\usepackage{booktabs}
\usepackage{makecell} 
\usepackage{pifont}   
\usepackage{multirow}
\usepackage{xcolor}
\newcommand{\cmark}{\ding{51}}

\usepackage{graphicx}
\usepackage{cuted}     
\usepackage{capt-of}

\usepackage{tcolorbox}
\tcbuselibrary{listings,skins,breakable}
\usepackage{enumitem}

\usepackage{caption}
\usepackage{cuted}
\usepackage{enumitem}

\newcommand{\mypara}[1]{\paragraph{\textbf{#1}}}

\usepackage{booktabs}
\usepackage{multirow}
\usepackage{adjustbox} 

\usepackage{booktabs}
\usepackage{adjustbox}
\usepackage{pifont}
\newcommand{\up}{\,$\uparrow$}
\newcommand{\down}{\,$\downarrow$}

\usepackage{booktabs,tabularx}

\begin{document}

\title{SceneOrchestra: Efficient Agentic 3D Scene Synthesis via Full Tool-Call Trajectory Generation} 

\titlerunning{SceneOrchestra}

\author{Yun He \and
Kelin Yu \and  Matthias Zwicker}

\authorrunning{Y. He~et al.}

\institute{University of Maryland, College Park, USA 
}

\maketitle
\vspace{-1.5em}

\begin{figure}[ht]
    \centering
    \begin{minipage}{0.96\linewidth}
        \centering
        \includegraphics[width=\linewidth]{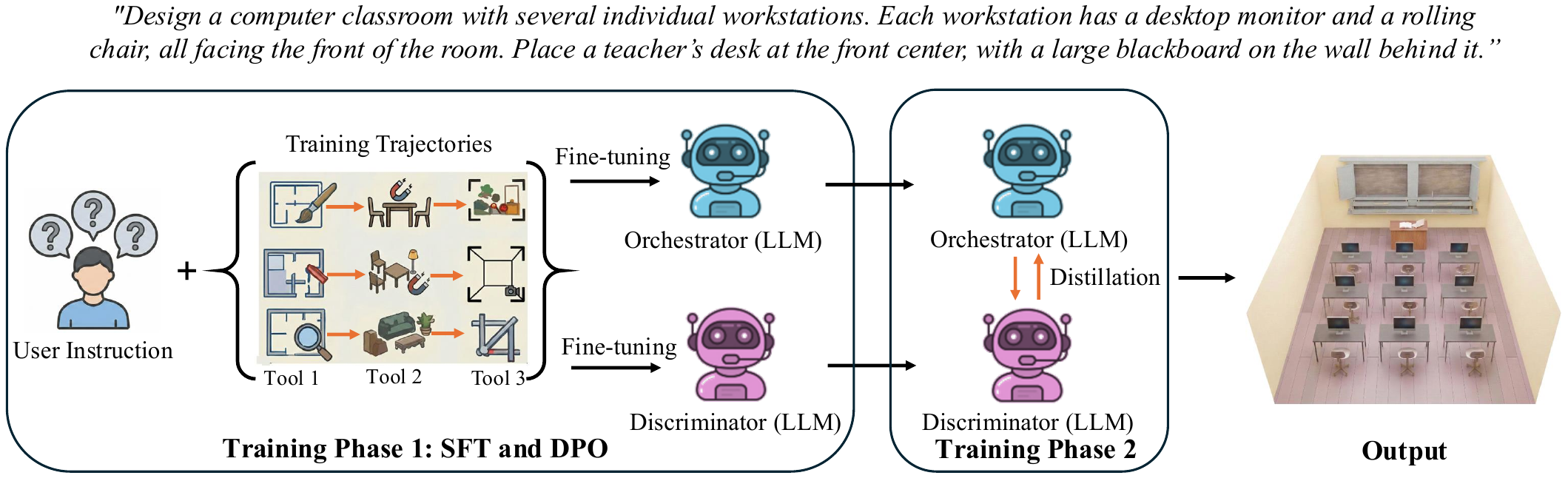}
        \captionof{figure}{SceneOrchestra is a trainable orchestration framework that outputs full tool-call trajectories to generate 3D scenes, optimized for efficiency and quality. Building on initial tool-call trajectories generated using an existing method, we train SceneOrchestra in two phases leveraging supervised fine tuning (SFT), direct preference optimization (DPO), and model distillation techniques. At inference time, SceneOrchestra outputs efficient tool call trajectories that generate high quality scenes in one shot.}
        \label{fig:teaser}
    \end{minipage}
\end{figure}

\vspace{-3em}

\begin{abstract}
Recent agentic frameworks for 3D scene synthesis have advanced realism and diversity by integrating heterogeneous generation and editing tools.
These tools are organized into workflows orchestrated by an off-the-shelf LLM.
Current approaches typically adopt an execute–review–reflect loop: at each step, the orchestrator executes a tool, renders intermediate results for review, and then decides on the tool and its parameters for the next step.
However, this design has two key limitations.
First, next-step tool selection and parameter configuration are driven by heuristic rules, which can lead to suboptimal execution flows, unnecessary tool invocations, degraded output quality, and increased runtime. Second, rendering and reviewing intermediate results after each step introduces additional latency. 
To address these issues, we propose SceneOrchestra, a trainable orchestration framework that optimizes the tool-call execution flow and eliminates the step-by-step review loop, improving both efficiency and output quality. 
SceneOrchestra consists of an orchestrator and a discriminator, which we fine-tune with a two-phase training strategy. 
In the first phase, the orchestrator learns context-aware tool selection and complete tool-call trajectory generation, while the discriminator is trained to assess the quality of full trajectories, enabling it to select the best trajectory from multiple candidates. 
In the second phase, we perform interleaved training, where the discriminator adapts to the orchestrator’s evolving trajectory distribution and distills its discriminative capability back into the orchestrator.
At inference, we only use the orchestrator to generate and execute full tool-call trajectories from instructions, without requiring the discriminator.
Extensive experiments show that our method achieves state-of-the-art scene quality compared to previous work.
\end{abstract}

\section{Introduction}
3D scene synthesis plays a crucial role in advancing virtual reality, embodied AI and robotic simulation \cite{wen20253d,zhang2019survey}.
As these fields evolve, natural language has emerged as a powerful interface for scene specification, offering intuitive control over complex generation processes \cite{raistrick2024infinigen,deitke2022,hollein2023text2room,ccelen2024design,sun2025layoutvlm,feng2023layoutgpt,yang2024holodeck,fu2024anyhome,ling2025scenethesis,gu2025artiscene,yang2025sceneweaver,he2022density,he2023grad}.
The ability to efficiently synthesize high-quality 3D scenes from textual descriptions not only democratizes content creation, but also opens new possibilities for personalized virtual experiences and adaptive simulation environments.

Previous 3D scene synthesis approaches can be roughly grouped into three categories: 1) rule-based methods \cite{deitke2022,raistrick2024infinigen} 
that rely on hand-crafted constraints for layout and object placement, 2) data-driven methods \cite{paschalidou2021atiss,tang2024diffuscene,
yang2024physcene,hollein2023text2room} that learn generative priors from curated scene datasets, and 3) LLM/VLM-powered approaches \cite{ccelen2024design,
sun2025layoutvlm,feng2023layoutgpt,yang2024holodeck,fu2024anyhome,ling2025scenethesis,gu2025artiscene} that leverage foundation models for semantic understanding. 
Each paradigm demonstrates specific strengths: rule-based methods ensure physical validity, data-driven approaches capture realistic scene distributions, and LLM/VLM-based methods offer flexible controllability. 
However, they remain fundamentally monolithic (only a single approach), unable to simultaneously achieve realism, fine-grained controllability, and physical reliability~\cite{yang2025sceneweaver}.

This limitation has motivated the development of SceneWeaver~\cite{yang2025sceneweaver}, an agentic framework that integrates multiple heterogeneous tools within a feedback-driven paradigm.
SceneWeaver iteratively invokes different tools to refine the scene and renders intermediate results after each step to guide subsequent tool selection.
While this iterative approach produces high-quality outputs compared to earlier techniques, it suffers from two critical issues.
1) Next-step tool selection is guided by heuristic rules.
Specifically, an off-the-shelf LLM~\cite{achiam2023gpt} acts as the orchestrator, selecting the next-step tool and its parameters based on manually summarized use cases.
This heuristic-driven strategy often leads to suboptimal execution flows, unnecessary tool invocations, degraded generation quality and increased runtime.
2) The feedback loop requires rendering and reviewing intermediate results after each step, which introduces additional overhead.
On average, SceneWeaver requires 1.5 hours per scene on a single A6000 GPU, severely limiting its practical scalability.

To address these limitations, we propose SceneOrchestra, a trainable tool orchestration framework that directly predicts complete tool-call trajectories from user instructions. This design eliminates heuristic, step-by-step tool selection and avoids the costly execute–review–reflect loop with repeated rendering and review. 
Specifically, SceneOrchestra comprises two fine-tuned LLMs \cite{yang2025qwen3}, an orchestrator and a discriminator. 
And we train the orchestrator and discriminator in two phases, 1) independent training and 2) interleaved training.

For independent training, we first collect tool-call trajectories as training data by running SceneWeaver~\cite{yang2025sceneweaver} multiple times per training instruction, producing diverse execution sequences. 
We train the orchestrator with a curriculum that combines stepwise and full trajectory-level supervision.
Stepwise training enables the orchestrator to perform local context-aware tool selection, while trajectory-level training guides the model to generate full trajectories for global planning. 
In addition, we use supervised fine-tuning (SFT) \cite{ouyang2022training} to learn basic prediction capabilities, and direct preference optimization (DPO) \cite{rafailov2023direct} to enable the orchestrator to self-evolve via preference learning. 
Meanwhile, we train the discriminator to rank trajectories according to generation quality and efficiency.

After the first phase, the trajectory distribution generated by the fine-tuned orchestrator deviates from the original distribution, since DPO pushes it towards higher quality trajectories. 
Hence, we employ interleaved training to adapt the discriminator to this updated distribution, and then distill its improved ranking ability back into the orchestrator.
This procedure alternates between two steps.
First, the orchestrator samples multiple trajectories for each user instruction in a training set.  
We then execute and rank the trajectories to fine-tune the discriminator.
Second, the orchestrator samples multiple trajectories for a new set of training instructions, and the updated discriminator ranks them without execution. We then use these rankings to construct preference pairs for trajectory-level DPO fine-tuning of the orchestrator, further improving its trajectory generation.
Repeating this cycle progressively distills the discriminator’s ranking capability back into the orchestrator. 
Finally, for inference we only invoke the orchestrator to generate  complete tool-call trajectories from user instructions and execute them end to end.

In summary, our contributions are as follows:
\begin{itemize}
\item We propose a trainable tool orchestration framework for efficient agentic 3D scene synthesis. 
It consists of an orchestrator that predicts complete tool-call trajectories from user instructions and a discriminator that selects the best trajectory among multiple candidates.

\item We introduce a two-phase training strategy. 
We first train the orchestrator and discriminator independently to obtain initial trajectory generation and ranking capabilities. 
We then perform interleaved training to adapt the discriminator to the orchestrator's evolving trajectory distribution and distill its assessment capability back into the orchestrator.

\item Extensive experiments show that our approach achieves state-of-the-art generation quality.
\end{itemize}

\section{Related Work}
\mypara{3D Indoor Scene Synthesis.}
Early works in 3D indoor scene synthesis mainly rely on procedural rules and handcrafted priors. Systems such as ProcTHOR~\cite{deitke2022} and Infinigen Indoors~\cite{raistrick2024infinigen} construct large-scale indoor scenes through rule-based procedural pipelines, which enable scalable scene creation. However, these approaches are highly dependent on manually designed heuristics, which offer limited flexibility for open-ended scene generation. With the availability of large 3D datasets, data-driven models have been proposed to learn scene layouts directly from data. 
Autoregressive and diffusion-based models, such as ATISS~\cite{paschalidou2021atiss}, DiffuScene~\cite{tang2024diffuscene}, and Text2Room~\cite{hollein2023text2room}, generate object arrangements by modeling spatial relationships between scene elements. 
Moreover, PhysScene~\cite{yang2024physcene} extends this paradigm to more realistic and interactive environments. 
Despite improved realism, these approaches typically still lack semantic controllability.

Recently, large language models (LLMs) and vision-language models (VLMs) have been explored for structured scene generation. Methods such as LayoutGPT~\cite{feng2023layoutgpt}, I-Design~\cite{ccelen2024design},  AnyHome~\cite{fu2024anyhome}, and Holodeck~\cite{yang2024holodeck} leverage LLMs to generate structured scene layouts from open-world textual instructions. Other works integrate language with visual feedback or optimization, including LayoutVLM~\cite{sun2025layoutvlm}, ArticScene~\cite{gu2025artiscene}, and SceneThesis~\cite{ling2025scenethesis}. While these methods improve semantic alignment and controllability, generating physically consistent indoor scenes remains challenging. To this end, recent work starts using agentic pipelines to ensure both semantic alignment and physical grounding. 

\vspace{-0.5em}
\mypara{Agentic Scene Synthesis.}
Scene synthesis relying on a single generative technique often struggles to simultaneously achieve realism, physical validity, and fine-grained controllability \cite{yang2025sceneweaver}. 
Hence, recent work \cite{yang2025sceneweaver,hu2024scenecraft,yin2026vision} is building on agentic workflows that integrate multiple tools to improve generation quality and control. 
Specifically, SceneWeaver \cite{yang2025sceneweaver} introduces an execute–review–reflect loop: 
at each step, it selects and executes a tool, renders the intermediate result for review, and then selects the next tool based on heuristics to iteratively refine the current output.
However, there are two limitations: 1) heuristic-based next-tool selection cannot guarantee an optimal execution flow, which may degrade quality and increase runtime; and 2) rendering and reviewing intermediate results at each step introduces additional latency. 
To address these issues, we propose a trainable orchestration framework that optimizes the execution flow and eliminates the step-by-step review loop, achieving higher quality in less time.

In addition, SceneSmith \cite{pfaff2026scenesmith} and SAGE \cite{xia2026sage} are concurrent works that also adopt agentic workflows. 
Compared to SceneWeaver \cite{yang2025sceneweaver}, they integrate more powerful tools, but their orchestration still relies on off-the-shelf LLMs. 
In contrast, we explicitly fine-tune an orchestrator for improved tool calling and global planning, distinguishing our approach  from these techniques.

 \vspace{-0.5em}
\mypara{Agentic Reinforcement Learning.}
Recent work has explored reinforcement learning (RL) techniques for training agentic systems that interact with environments through tool use~\cite{rltooluse, react,li2024league++, he2025langdrivectrl}. Offline RL methods learn from pre-collected datasets without requiring additional environment interaction~\cite{levine2020offline, fujimoto2019off, kumar2020conservative, kostrikov2021offline}, enabling safe learning from logged data. In contrast, online RL approaches continuously interact with environments to collect new data~\cite{schulman2017proximal, hafner2020mastering, schrittwieser2020mastering, shao2024deepseekmath}, allowing policies to adapt based on real-time feedback. 
In the context of agentic systems~\cite{agenticrl}, offline training often relies on demonstrations and preference data. Supervised fine-tuning (SFT)~\cite{ouyang2022training,minaee2024large} initializes the policy by imitating expert demonstrations, while Direct Preference Optimization (DPO)~\cite{rafailov2023direct} further refines it using pre-collected preference data.
More recent online methods such as GRPO~\cite{shao2024deepseekmath} perform real-time policy optimization by continuously collecting trajectories and updating policies based on immediate environment feedback. However, such online methods are impractical in our setting, where executing a single rollout takes over one hour, making real-time policy updates infeasible. Therefore, we adopt SFT and DPO in our approach.

\section{Method}

\begin{figure*}[t!]
    \centering
    \includegraphics[width=\linewidth]{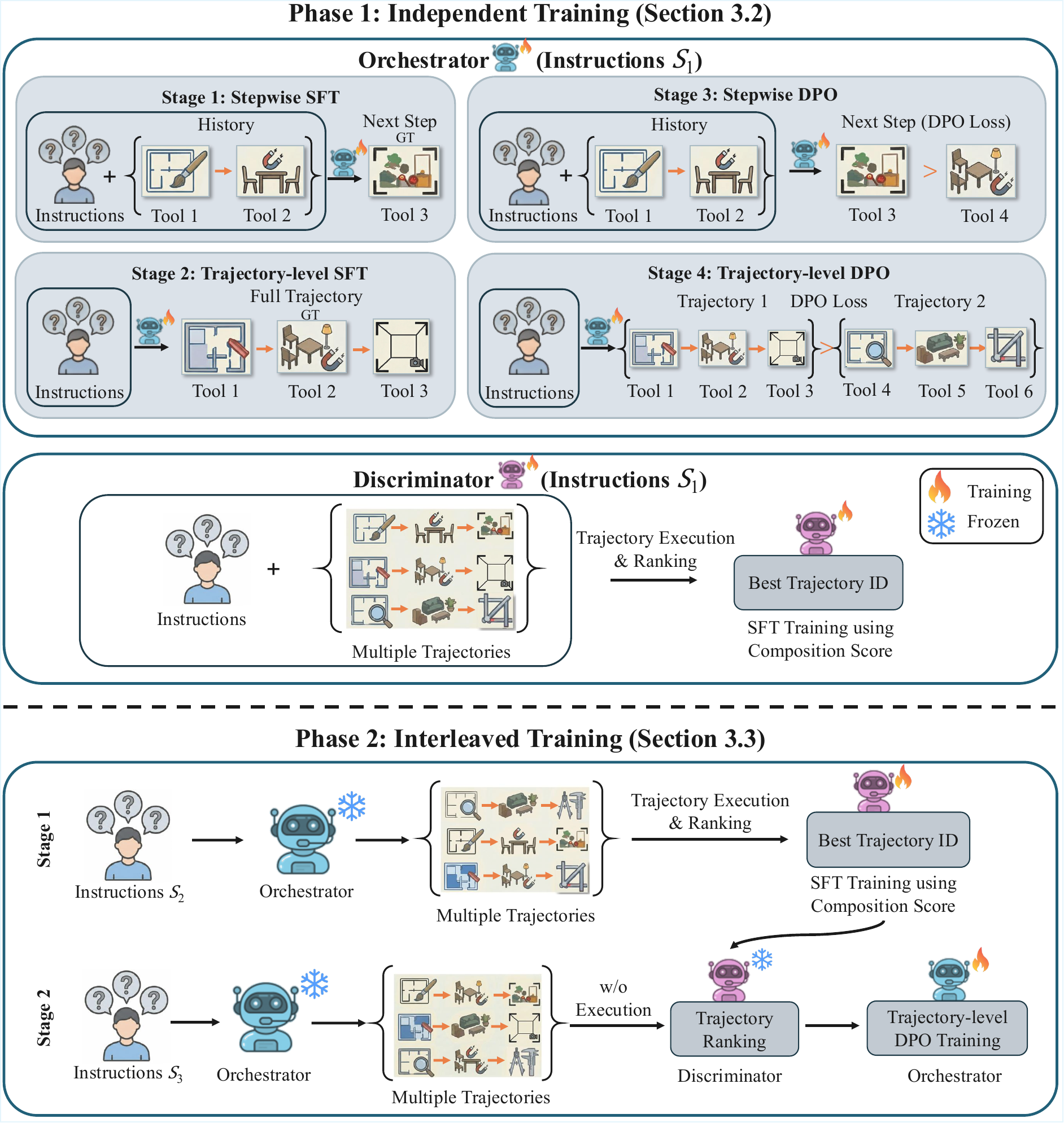}
    \vspace{-1.2em}
    \caption{\textbf{Training Paradigm.} Our training consists of two phases. In the independent training phase, we train the orchestrator and discriminator separately to develop their initial capabilities. 
    The orchestrator undergoes four stages: 1) stepwise SFT, 2) trajectory-level SFT, 3) stepwise DPO, and 4) trajectory-level DPO, learning context-aware tool selection and complete trajectory generation. 
    Meanwhile, the discriminator learns to select the optimal trajectory from multiple candidates. 
    In the interleaved training phase, we adapt the discriminator to the orchestrator's evolving trajectory distribution, then distill this discriminative capability back into the orchestrator. 
    This enables inference with only the orchestrator, eliminating the need for the discriminator.
    }
    \label{fig:pipeline}
    \vspace{-1.6em}
\end{figure*}

SceneOrchestra consists of two fine-tuned LLM components based on Qwen3~\cite{yang2025qwen3}: an orchestrator and a discriminator.
The orchestrator learns to generate complete tool-call trajectories from instructions, while the discriminator learns to select the optimal trajectory from multiple candidates. 
Our training proceeds in two phases, as shown in Fig. \ref{fig:pipeline}. 
In the first phase, we train both components independently to develop initial trajectory generation and assessment capabilities. 
In the second phase, we employ interleaved training that adapts the discriminator to the orchestrator's evolving trajectory distribution and distills its assessment capability back into the orchestrator.
During inference, the orchestrator directly generates a full tool-call trajectory and executes it end-to-end within SceneWeaver's \cite{yang2025sceneweaver} framework, without requiring the discriminator.

\subsection{Preliminaries}
We employ two complementary techniques to fine-tune our models in the first phase: supervised fine-tuning (SFT) and direct preference optimization (DPO).

\mypara{Supervised Fine-Tuning (SFT).}
SFT \cite{ouyang2022training,minaee2024large} adapts a pretrained language model (we use Qwen3 \cite{yang2025qwen3}) to a target task by learning from demonstrations. 
Given training data $\mathcal{D}_{\text{SFT}} = \{(x_i, y_i)\}_{i=1}^N$ containing input-output 
pairs, SFT maximizes the likelihood of generating correct outputs:
\begin{equation}
\mathcal{L}_{\text{SFT}}(\theta) = -\mathbb{E}_{(x,y) \sim \mathcal{D}_{\text{SFT}}} 
\left[ \log \pi_\theta(y | x) \right]
\label{eq:sft}
\end{equation}
where $\pi_\theta$ denotes the model parameterized by $\theta$. This 
establishes basic task-solving capability and output format alignment.

\mypara{Direct Preference Optimization (DPO).}
DPO~\cite{rafailov2023direct} improves upon SFT by learning from comparative feedback. 
Given a preference dataset $\mathcal{D}_{\text{DPO}}$ containing triplets $(x_i, y_i^w, y_i^l)$, where output $y_i^w$ is preferred over $y_i^l$,
DPO optimizes:
\begin{equation}
\mathcal{L}_{\text{DPO}}(\theta) = -\mathbb{E}_{(x, y^w, y^l) \sim \mathcal{D}_{\text{DPO}}} 
\left[ \log \sigma \left( \beta \log \frac{\pi_\theta(y^w | x)}{\pi_{\text{ref}}(y^w | x)} 
- \beta \log \frac{\pi_\theta(y^l | x)}{\pi_{\text{ref}}(y^l | x)} \right) \right]
\label{eq:dpo}
\end{equation}
where $\pi_{\text{ref}}$ is the reference model, $\beta$ controls the sharpness of the preference signal, and $\sigma$ is the sigmoid function. 
This encourages the model to favor higher-quality outputs $y_i^w$ while staying close to the reference model.

\subsection{Independent Training}
\vspace{-0.3em}
To obtain complete tool-call trajectories for training, we first construct a training instruction set $\mathcal{S}_1$. 
We then run SceneWeaver~\cite{yang2025sceneweaver} multiple times for each instruction to generate diverse tool trajectory rollouts.
For step $t$ in each rollout $i$, we record its quality score $Q^{(i,t)}$ and cumulative runtime $T^{(i,t)}$.

\mypara{Quality Score.} 
For quality assessment, we adopt the evaluation metrics from 
SceneWeaver~\cite{yang2025sceneweaver}, which consist of physical and 
visual-semantic components. 
We use GPT-4.1 \cite{openai2025gpt41} as the evaluator to compute these scores.
Physical metrics include the number of objects 
in the scene ($N_{\text{obj}}^{(i,t)}$), number of out-of-boundary objects 
($N_{\text{ob}}^{(i,t)}$), and number of collided object pairs ($N_{\text{col}}^{(i,t)}$). 
The physical score is computed as 
$Q_{\text{phy}}^{(i,t)} = N_{\text{obj}}^{(i,t)} - \alpha(N_{\text{ob}}^{(i,t)} + N_{\text{col}}^{(i,t)})$, 
where $\alpha$ is a penalty coefficient. 
Visual metrics evaluate visual 
realism ($S_{\text{real}}^{(i,t)}$), functionality ($S_{\text{func}}^{(i,t)}$), 
layout correctness ($S_{\text{lay}}^{(i,t)}$), and scene completeness ($S_{\text{comp}}^{(i,t)}$). 
The visual score is 
$Q_{\text{vis}}^{(i,t)} = (S_{\text{real}}^{(i,t)} + S_{\text{func}}^{(i,t)} + S_{\text{lay}}^{(i,t)} + S_{\text{comp}}^{(i,t)})/4$. 
The overall quality score combines both components:
\begin{equation}
Q^{(i,t)} = \lambda Q_{\text{phy}}^{(i,t)} + Q_{\text{vis}}^{(i,t)}
\label{eq:quality}
\end{equation}
where $\lambda$ balances physical and visual quality.

\mypara{Composition Score.} 
We use a composition score to jointly consider generation quality and efficiency:
\begin{equation}
C^{(i,t)} = Q^{(i,t)} - \gamma T^{(i,t)}
\label{eq:composition}
\end{equation}
where $\gamma$ is a coefficient that balances quality and time cost.

\subsubsection{Orchestrator.}
In this phase, the orchestrator is trained in four stages: 1) stepwise SFT, 2) trajectory-level SFT, 3) stepwise DPO, and 4) trajectory-level DPO. 
Each stage initializes from the checkpoint of the previous stage, which also serves as the reference model for DPO training.
Stepwise training predicts the next tool call given the instruction and the execution history, teaching the model context-aware tool selection. 
Trajectory-level training, in contrast, requires the model to generate a complete trajectory conditioned only on the instruction. 
Since language models generate outputs token-by-token, stepwise training naturally provides a foundation for trajectory-level training.
In addition, we first use SFT to establish the desired output format and basic prediction ability, and then apply DPO to further improve trajectory generation through preference learning.
All SFT and DPO stages consistently employ their respective optimization objectives discussed above (Equations~\ref{eq:sft} and~\ref{eq:dpo}).

In the following, we detail the training data construction process for each stage, with the goal of curating a high-quality data set from the initial tool call trajectories obtained using SceneWeaver~\cite{yang2025sceneweaver}.

\mypara{Stepwise SFT.}
In this stage, we focus on teaching the orchestrator 
basic context-aware tool selection ability. For each rollout $i$, we identify steps where the composition score increases significantly.
Specifically, consecutive steps where $C^{(i,t)} - C^{(i,t-1)} > \tau_1$, with $\tau_1$ being a threshold. 
We then construct stepwise SFT training pairs $\{x_s, y_s\}$, where $x_s$ consists of the instruction, available tool list, and execution history, and $y_s$ is the next tool call.

\mypara{Trajectory-level SFT.}
In this stage, we aim to teach the orchestrator 
complete trajectory prediction ability and align its output format for inference. 
For each rollout $i$, we randomly truncate the trajectory at step $t$ and evaluate its quality using the composition score $C^{(i,t)}$ at that step. 
If $C^{(i,t)} > \tau_2$, where $\tau_2$ is a predefined threshold, we consider it a high-quality trajectory and keep it. 
We then construct trajectory-level SFT training 
pairs $\{x_t, y_t\}$, where $x_t$ contains the instruction and available tool list, and $y_t$ is the complete trajectory up to step $t$.

\vspace{-0.5em}
\mypara{Stepwise DPO.}
For this stage, we aim to enhance the orchestrator's next tool selection ability by comparing alternative tool choices under the same execution history. 
For each rollout $i$, we first identify steps with 
significant composition score changes—consecutive steps where $|C^{(i,t)} - C^{(i,t-1)}| > \tau_1$. 
Since the orchestrator has acquired basic prediction capability from the previous two stages, we 
feed the execution history of these selected steps to the fine-tuned orchestrator to predict alternative tools and their parameters. 
We then execute the predicted tool and obtain its composition score $C_{\text{pred}}^{(i,t)}$, which we compare with the original tool's score $C_{\text{orig}}^{(i,t)}$. 
If the score difference exceeds threshold $\tau_3$ ($|C_{\text{pred}}^{(i,t)} - C_{\text{orig}}^{(i,t)}| > \tau_3$), we construct a DPO training triplet $(x_s, y_s^{w}, y_s^{l})$, where $x_s$ 
contains the instruction, available tool list, and execution history; $y_s^{w}$ is the chosen tool (with higher score); and $y_s^{l}$ is the rejected tool (with lower score).

\vspace{-0.5em}
\mypara{Trajectory-level DPO.}
To improve the orchestrator's ability to 
generate complete trajectories, we leverage pairwise comparisons between different execution sequences derived from the same instruction. 
Specifically, for each instruction, we obtain two rollouts $i$ and $j$ through random sampling, and truncate each at independently chosen steps $t_i$ and $t_j$. 
The resulting trajectories are evaluated using their composition scores $C^{(i,t_i)}$ and 
$C^{(j,t_j)}$. 
When the score gap is larger than threshold $\tau_4$ 
($|C^{(i,t_i)} - C^{(j,t_j)}| > \tau_4$), we form a DPO training triplet $(x_t, y_t^{w}, y_t^{l})$, where $x_t$ 
specifies the instruction and tool list, $y_t^{w}$ represents the superior trajectory (higher score), and $y_t^{l}$ represents the inferior one (lower score).

\vspace{-0.5em}
\subsubsection{Discriminator.}
The orchestrator learns a conditional distribution over 
high-quality trajectories and samples from it during inference.
However, this does not guarantee optimal trajectory 
selection, as the orchestrator lacks explicit trajectory classification or ranking capability. 
To address this limitation, inspired by 
GAN~\cite{goodfellow2020generative}, we introduce a discriminator that we train to perform trajectory-level assessment through SFT.
Specifically, for each instruction, we sample multiple truncated trajectories from its rollouts, and compute their composition scores as in Eq.~\ref{eq:composition}. 
The trajectory with the highest score is designated as the best. 
We then construct discriminator SFT training pairs $\{x_d, y_d\}$, where $x_d$ contains the instruction, available tool list and a set of candidate trajectories, and $y_d$
is the index of the best trajectory.

\subsection{Interleaved Training}
After independent training, the orchestrator and discriminator have developed basic trajectory generation and assessment capabilities. 
However, the trajectory distribution generated by the orchestrator evolved since we employed DPO to improve  over the original distribution. 
Specifically, the fine-tuned orchestrator induces a trajectory distribution that differs from the original SceneWeaver \cite{yang2025sceneweaver}.
We address this distribution shift through interleaved training, which adapts the discriminator to the orchestrator's evolving generation distribution, and then distills its updated discriminative capability back into the orchestrator to further improve trajectory quality.

Interleaved training alternates between two stages shown in Fig.~\ref{fig:pipeline}.
At first, the orchestrator samples and executes multiple trajectories for each training instruction, denoted $\mathcal{S}_2$ in Fig.~\ref{fig:pipeline}. 
We then select the best trajectory based on composition scores (Eq.~\ref{eq:composition}), construct discriminator training 
data $\{x_d, y_d\}$ as in the first phase, and fine-tune the discriminator.
Second, the orchestrator generates multiple candidates for each new instruction, denoted $\mathcal{S}_3$ in Fig.~\ref{fig:pipeline}. 
Then the updated discriminator directly identifies the best trajectory without time-consuming execution. 
This provides preference signals $(x_t, y_t^{w}, y_t^{l})$ that we use for trajectory-level DPO training of the orchestrator. While these two stages could be repeated in principle, we perform one interleaved training phase in practice.

Finally, at inference time the discriminator is no longer needed and the orchestrator generates complete trajectories and executes them end to end.

\section{Experiments}

\subsection{Experiment Setup}
\mypara{Settings.}
Following SceneWeaver~\cite{yang2025sceneweaver}, we use the template-based prompt "Design me a <room\_type>" as test instructions. 
We evaluate on 10 room types: 5 unseen types (bedroom, living room, kitchen, gym, restaurant) that never appear in training, and 5 seen types (bathroom, children room, meeting room, office, waiting room) that appear in training but with different instructions. 
Specifically, training instructions provide detailed 
specifications of object counts, positions, layouts, etc. 
While test instructions use only the generic template. 
For each instruction, we generate 5 scenes and report the average metrics.
In addition, we also use complex instructions to test model's fine-grained control ability.

\begin{figure*}[t!]
    \centering
    \includegraphics[width=\linewidth]{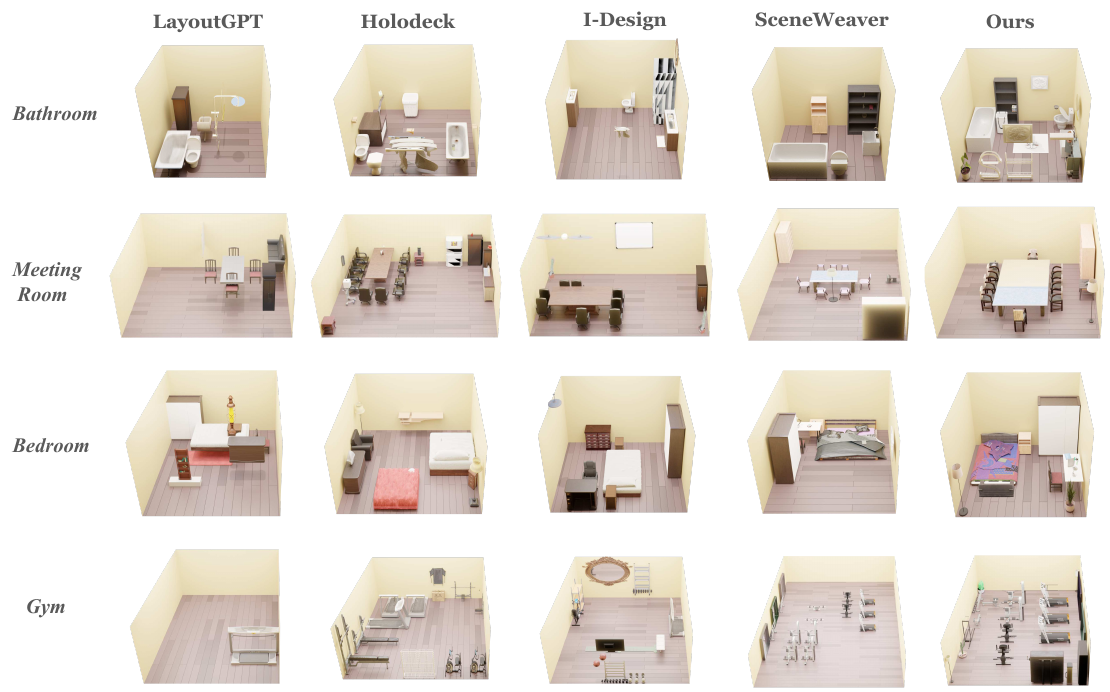}
    \caption{\textbf{Qualitative comparison with baselines} on both seen (bathroom, meeting room) and unseen (bedroom, gym) room types. Our method achieves higher visual realism, more plausible layouts, and richer details.
    }
    \label{fig:base_quality}
    \vspace{-1.5em}
\end{figure*}

\mypara{Baselines.}
Following SceneWeaver~\cite{yang2025sceneweaver}, we 
compare against state-of-the-art LLM-powered 3D scene synthesis methods that support open-vocabulary queries: LayoutGPT~\cite{feng2023layoutgpt}, Holodeck~\cite{yang2024holodeck}, and I-Design~\cite{ccelen2024design}. 
We also include SceneWeaver itself as a baseline. 
Note that the difference between our method and SceneWeaver lies solely in the tool-call trajectory generation—we use the same tool implementations from SceneWeaver's framework.

\mypara{Metrics.}
For a fair comparison, we strictly follow the same evaluation protocol as SceneWeaver~\cite{yang2025sceneweaver}. The metrics are organized into two categories: physical metrics and visual-semantic metrics. Physical metrics are directly computed from the generated scene geometry, including the average object count (\#Obj), number of out-of-boundary objects (\#OB), and collided object pairs (\#CN). For visual-semantic metrics, each scene is evaluated by providing GPT-4.1~\cite{openai2025gpt41} with the instruction and a top-down rendered view. These metrics assess perceptual quality across four dimensions: visual realism (Real.), functionality (Func.), layout correctness (Lay.), and scene completeness (Comp.).

\mypara{Implementation Details.}
We use Qwen3-4B-Instruct-2507~\cite{yang2025qwen3} as the base model for both the orchestrator and the discriminator. All SFT and DPO training is conducted using the LLaMAFactory framework~\cite{zheng2024llamafactory}, and we adopt LoRA~\cite{hu2022lora} for parameter-efficient fine-tuning. We use GPT-4~\cite{achiam2023gpt} to generate the training instructions, which are then manually checked and filtered. These instructions provide detailed descriptions of indoor scenes, including object counts, object positions, layouts, and other scene-level constraints. We further apply instruction rephrasing as a form of data augmentation during training.

\subsection{Experimental Results}

\begin{table*}[t]
  \centering
  \caption{\textbf{Quantitative comparison with baselines} on seen and unseen room types. 
  We use the template-based instruction "Design me a <room\_type>". For each room type, all methods are executed five times and the results are averaged. Overall, our method achieves the best performance on nearly all metrics.
}
  \vspace{-0.5em}
  \label{tab:open_vocab_10rooms}

  \setlength{\tabcolsep}{3.5pt}
  \renewcommand{\arraystretch}{1.05}

  \begin{adjustbox}{max width=0.75\linewidth}
  \begin{tabular}{l ccccccc ccccccc}
    \toprule
    \multirow{2}{*}{\textbf{Method}} &
    \multicolumn{7}{c}{\textbf{Bedroom (Unseen)}} &
    \multicolumn{7}{c}{\textbf{Living Room (Unseen)}} \\
    \cmidrule(lr){2-8}\cmidrule(lr){9-15}
    & \#Obj\up & \#OB\down & \#CN\down & Real.\up & Func.\up & Lay.\up & Comp.\up
    & \#Obj\up & \#OB\down & \#CN\down & Real.\up & Func.\up & Lay.\up & Comp.\up \\
    \midrule
    LayoutGPT \cite{feng2023layoutgpt}      & 6.8 & 1.2 & 1.6 & 7.2 & 8.4 & 6.2 & 4.2  & 8.4 & 0.8 & 2.0 & 6.4 & 6.6 & 5.4 & 3.8 \\
    Holodeck \cite{yang2024holodeck}        & \textbf{28.8} & 0.0 & 34.4 & 8.0 & 8.6 & 6.4 & 6.0  & \textbf{32.0} & 0.0 & 10.2 & 8.8 & 9.2 & 7.4 & 8.0 \\
    I-Design \cite{ccelen2024design}        & 8.8 & 0.0 & 0.0 & 7.8 & 9.0 & 7.2 & 5.2  & 9.0 & 0.0 & 0.0 & 7.4 & 8.2 & 6.6 & 4.8 \\
    SceneWeaver \cite{yang2025sceneweaver}  & 13.6 & 0.0 & 0.0 & 8.8 & 9.0 & 7.4 & 6.6  & 16.0 & 0.0 & 0.0 & 8.8 & 8.6 & 7.8 & 7.4 \\
    \textbf{Ours}                                   & 15.4 & \textbf{0.0} & \textbf{0.0} & \textbf{9.0} & \textbf{9.2} & \textbf{8.0} & \textbf{8.2}  & 18.6 & \textbf{0.0} & \textbf{0.0} & \textbf{9.2} & \textbf{9.4} & \textbf{8.0} & \textbf{8.6} \\
    \bottomrule
  \end{tabular}
  \end{adjustbox}

  \vspace{0.35em}

  \begin{adjustbox}{max width=\linewidth}
  \begin{tabular}{l ccccccc ccccccc ccccccc}
    \toprule
    \multirow{2}{*}{\textbf{Method}} &
    \multicolumn{7}{c}{\textbf{Kitchen (Unseen)}} &
    \multicolumn{7}{c}{\textbf{Gym (Unseen)}} &
    \multicolumn{7}{c}{\textbf{Restaurant (Unseen)}} \\
    \cmidrule(lr){2-8}\cmidrule(lr){9-15}\cmidrule(lr){16-22}
    & \#Obj\up & \#OB\down & \#CN\down & Real.\up & Func.\up & Lay.\up & Comp.\up
    & \#Obj\up & \#OB\down & \#CN\down & Real.\up & Func.\up & Lay.\up & Comp.\up
    & \#Obj\up & \#OB\down & \#CN\down & Real.\up & Func.\up & Lay.\up & Comp.\up \\
    \midrule
    LayoutGPT \cite{feng2023layoutgpt}      & 8.6 & 1.6 & 2.0 & 5.8 & 6.4 & 5.0 & 3.8  & 4.4 & 0.8 & 0.0 & 5.4 & 4.4 & 5.8 & 2.4  & 9.4 & 1.4 & 2.2 & 3.6 & 3.4 & 4.8 & 2.4 \\
    Holodeck \cite{yang2024holodeck}        & 19.6 & 0.0 & 3.2 & 6.6 & 5.8 & 6.0 & 4.0  & 24.2 & 0.0 & 4.8 & 9.2 & \textbf{9.4} & 7.2 & 6.6  & 31.0 & 0.0 & 7.8 & 4.8 & 3.6 & 4.2 & 3.4 \\
    I-Design \cite{ccelen2024design}        & 10.4 & 0.0 & 0.0 & 6.2 & 6.6 & 5.6 & 3.4  & 11.8 & 0.0 & 0.8 & 8.4 & 8.8 & 7.2 & 6.0  & 27.2 & 0.0 & 0.0 & 4.8 & 4.6 & 5.0 & 3.2 \\
    SceneWeaver \cite{yang2025sceneweaver}  & 32.4 & 0.0 & 0.0 & 9.0 & 8.8 & 7.4 & 7.8  & 21.2 & 0.0 & 0.0 & 8.8 & 9.2 & 7.2 & 7.6  & 71.4 & 0.0 & 1.4 & 8.2 & 7.4 & 6.8 & 6.0 \\
    \textbf{Ours}                                   & \textbf{34.8} & \textbf{0.0} & \textbf{0.0} & \textbf{9.2} & \textbf{9.2} & \textbf{7.6} & \textbf{8.0}  & \textbf{27.6} & \textbf{0.0} & \textbf{0.0} & \textbf{9.2} & 9.2 & \textbf{8.0} & \textbf{7.8}  & \textbf{78.2} & \textbf{0.0} & \textbf{0.0} & \textbf{8.6} & \textbf{8.0} & \textbf{7.8} & \textbf{7.4} \\
    \bottomrule
  \end{tabular}
  \end{adjustbox}

  \vspace{0.35em}

  \begin{adjustbox}{max width=\linewidth}
  \begin{tabular}{l ccccccc ccccccc ccccccc}
    \toprule
    \multirow{2}{*}{\textbf{Method}} &
    \multicolumn{7}{c}{\textbf{Bathroom (Seen)}} &
    \multicolumn{7}{c}{\textbf{Children Room (Seen)}} &
    \multicolumn{7}{c}{\textbf{Meeting Room (Seen)}} \\
    \cmidrule(lr){2-8}\cmidrule(lr){9-15}\cmidrule(lr){16-22}
    & \#Obj\up & \#OB\down & \#CN\down & Real.\up & Func.\up & Lay.\up & Comp.\up
    & \#Obj\up & \#OB\down & \#CN\down & Real.\up & Func.\up & Lay.\up & Comp.\up
    & \#Obj\up & \#OB\down & \#CN\down & Real.\up & Func.\up & Lay.\up & Comp.\up \\
    \midrule
    LayoutGPT \cite{feng2023layoutgpt}      & 7.4 & 1.2 & 0.8 & 8.2 & 9.0 & 6.8 & 5.4  & 8.6 & 1.4 & 1.0 & 4.8 & 5.0 & 5.2 & 3.0  & 8.4 & 1.2 & 1.6 & 5.4 & 5.2 & 5.0 & 3.0 \\
    Holodeck \cite{yang2024holodeck}        & 12.4 & 0.0 & 1.4 & 8.0 & 7.4 & 6.6 & 5.4  & 17.8 & 0.0 & 3.6 & 7.8 & 8.4 & 6.6 & 6.0  & 29.2 & 0.0 & 3.4 & 8.8 & 9.0 & 7.4 & 7.2 \\
    I-Design \cite{ccelen2024design}        & 8.4 & 0.0 & 0.0 & 7.6 & 8.4 & 7.0 & 4.8  & 8.8 & 0.0 & 0.0 & 6.8 & 7.8 & 6.0 & 4.8  & 17.2 & 2.6 & 0.4 & 6.6 & 6.8 & 5.8 & 4.4 \\
    SceneWeaver \cite{yang2025sceneweaver}  & 12.0 & 0.0 & 0.0 & 9.0 & 9.2 & 7.4 & \textbf{8.2}  & 16.0 & 0.0 & 1.0 & 9.2 & 9.2 & 7.2 & 8.4  & 26.4 & 0.0 & 0.0 & 8.6 & 8.2 & 7.0 & 7.6 \\
    \textbf{Ours}                                   & \textbf{14.2} & \textbf{0.0} & \textbf{0.0} & \textbf{9.2} & \textbf{9.4} & \textbf{8.0} & 8.0  & \textbf{19.6} & \textbf{0.0} & \textbf{0.0} & \textbf{9.4} & \textbf{9.2} & \textbf{7.8} & \textbf{8.6}  & \textbf{31.2} & \textbf{0.0} & \textbf{0.0} & \textbf{9.0} & \textbf{9.0} & \textbf{8.2} & \textbf{8.4} \\
    \bottomrule
  \end{tabular}
  \end{adjustbox}

  \vspace{0.35em}

  \begin{adjustbox}{max width=\linewidth}
  \begin{tabular}{l ccccccc ccccccc ccccccc}
    \toprule
    \multirow{2}{*}{\textbf{Method}} &
    \multicolumn{7}{c}{\textbf{Office (Seen)}} &
    \multicolumn{7}{c}{\textbf{Waiting Room (Seen)}} &
    \multicolumn{7}{c}{\textbf{Average}} \\
    \cmidrule(lr){2-8}\cmidrule(lr){9-15}\cmidrule(lr){16-22}
    & \#Obj\up & \#OB\down & \#CN\down & Real.\up & Func.\up & Lay.\up & Comp.\up
    & \#Obj\up & \#OB\down & \#CN\down & Real.\up & Func.\up & Lay.\up & Comp.\up
    & \#Obj\up & \#OB\down & \#CN\down & Real.\up & Func.\up & Lay.\up & Comp.\up \\
    \midrule
    LayoutGPT \cite{feng2023layoutgpt}      & 8.4 & 0.6 & 0.6 & 6.6 & 7.2 & 5.6 & 3.8  & 7.4 & 1.0 & 0.6 & 6.8 & 6.8 & 5.6 & 3.8  &  7.8  & 1.1 & 1.2 & 6.0 & 6.2 & 5.5 & 3.6 \\
    Holodeck \cite{yang2024holodeck}        & 36.2 & 0.0 & 10.8 & 7.8 & 7.2 & 5.6 & 5.6 & 30.8 & 0.0 & 6.6 & 8.4 & 9.0 & 6.6 & 6.0  & 26.2 & 0.0 & 8.6 & 7.8 & 7.8 & 6.4 & 5.8 \\
    I-Design \cite{ccelen2024design}        & 9.8 & 0.0 & 0.0 & 7.2 & 8.4 & 6.6 & 4.6  & 11.4 & 0.0 & 0.0 & 6.6 & 7.2 & 5.8 & 4.2  & 12.3 & 0.3 & 0.1 & 6.9 & 7.6 & 6.3 & 4.5 \\
    SceneWeaver \cite{yang2025sceneweaver}  & 35.2 & 0.0 & 0.0 & 8.8 & 8.0 & 6.8 & 7.4  & 26.2 & 0.0 & 0.0 & 8.8 & 9.0 & 7.4 & 7.6  & 27.0 & 0.0 & 0.2 & 8.8 & 8.7 & 7.2 & 7.5 \\
    \textbf{Ours}                                   & \textbf{38.6} & \textbf{0.0} & \textbf{0.0} & \textbf{9.2} & \textbf{8.8} & \textbf{7.6} & \textbf{8.2}  & \textbf{32.6} & \textbf{0.0} & \textbf{0.0} & \textbf{9.2} & \textbf{9.2} & \textbf{8.0} & \textbf{8.0}  & \textbf{31.1} & \textbf{0.0} & \textbf{0.0} & \textbf{9.1} & \textbf{9.1} & \textbf{7.9} & \textbf{8.1} \\
    \bottomrule
  \end{tabular}
  \end{adjustbox}

  \vspace{-1.0em}
\end{table*}

We first report quantitative comparisons with baselines in Tab. \ref{tab:open_vocab_10rooms}. The evaluated room types fall into two categories: unseen room types that never appear in the training instructions, and seen room types that appear in training, but with different instructions. 
Overall, our method achieves the best performance on almost all metrics for both seen and unseen room types.
Specifically, although Holodeck \cite{yang2024holodeck} sometimes generates scenes with more objects than ours, it often suffers from collisions, as its physical and geometric constraints are weak and the placement is closer to random. 
Our method also significantly outperforms SceneWeaver, despite using the same set of tools. 
This demonstrates that our fine-tuned orchestrator learns more effective tool selection and complete trajectory generation, thereby improving the overall tool-call execution flow.

We further provide qualitative comparisons in Fig.~\ref{fig:base_quality}. 
LayoutGPT~\cite{feng2023layoutgpt} produces poor physical layouts, such as tables and chairs intersecting each other, while Holodeck~\cite{yang2024holodeck} and I-Design~\cite{ccelen2024design} often yield unrealistic arrangements. 
SceneWeaver~\cite{yang2025sceneweaver} produces generally plausible results but still exhibits subtle inconsistencies, such as a toilet placed next to a bathtub in a bathroom or a floor lamp positioned awkwardly beside a chair in a meeting room. 
In contrast, our method generates layouts that are consistently coherent and realistic.

\vspace{-1em}
\begin{table*}[htbp]
  \centering
  \caption{\textbf{Average generation time comparison (minutes).} We report the average generation time across all room types. All methods run on a single A6000 GPU. Monolithic methods are much faster than agentic approaches, since agentic methods always invoke multiple tools. By optimizing the execution flow and removing the step-by-step review process, our method substantially reduces the runtime of agentic generation.}
  \vspace{-0.5em}
  \label{tab:generation_time}
  \setlength{\tabcolsep}{8pt}
  \renewcommand{\arraystretch}{1.05}
  \begin{adjustbox}{width=0.75\textwidth}
  \begin{tabular}{lccccc}
    \toprule
    &
    \textbf{LayoutGPT}~\cite{feng2023layoutgpt} &
    \textbf{Holodeck}~\cite{yang2024holodeck} &
    \textbf{I-Design}~\cite{ccelen2024design} &
    \textbf{SceneWeaver}~\cite{yang2025sceneweaver} &
    \textbf{Ours} \\
    \midrule
    \textbf{Time (min) $\downarrow$} & 2.4 & 12.4 & 6.1 & 91.0 & 39.7 \\
    \bottomrule
  \end{tabular}
  \end{adjustbox}

  \vspace{-1.5em}
\end{table*}

Beyond generation quality, we also report the runtime of different methods in Tab. \ref{tab:generation_time} (Detailed generation times for each room type are provided in the Tab.~\ref{tab:time_for_each_roomtype}). 
LayoutGPT \cite{feng2023layoutgpt}, Holodeck \cite{yang2024holodeck}, and I-Design \cite{ccelen2024design} are relatively fast because they are monolithic approaches.
However, this efficiency comes at the cost of lower generation quality. 
SceneWeaver \cite{yang2025sceneweaver} achieves higher-quality results by integrating multiple tools, but it is much slower. 
In contrast, our method optimizes the execution flow and removes the step-by-step review process, significantly reducing runtime while further improving generation quality.

\begin{figure*}[t]
    \centering
    \includegraphics[width=\linewidth]{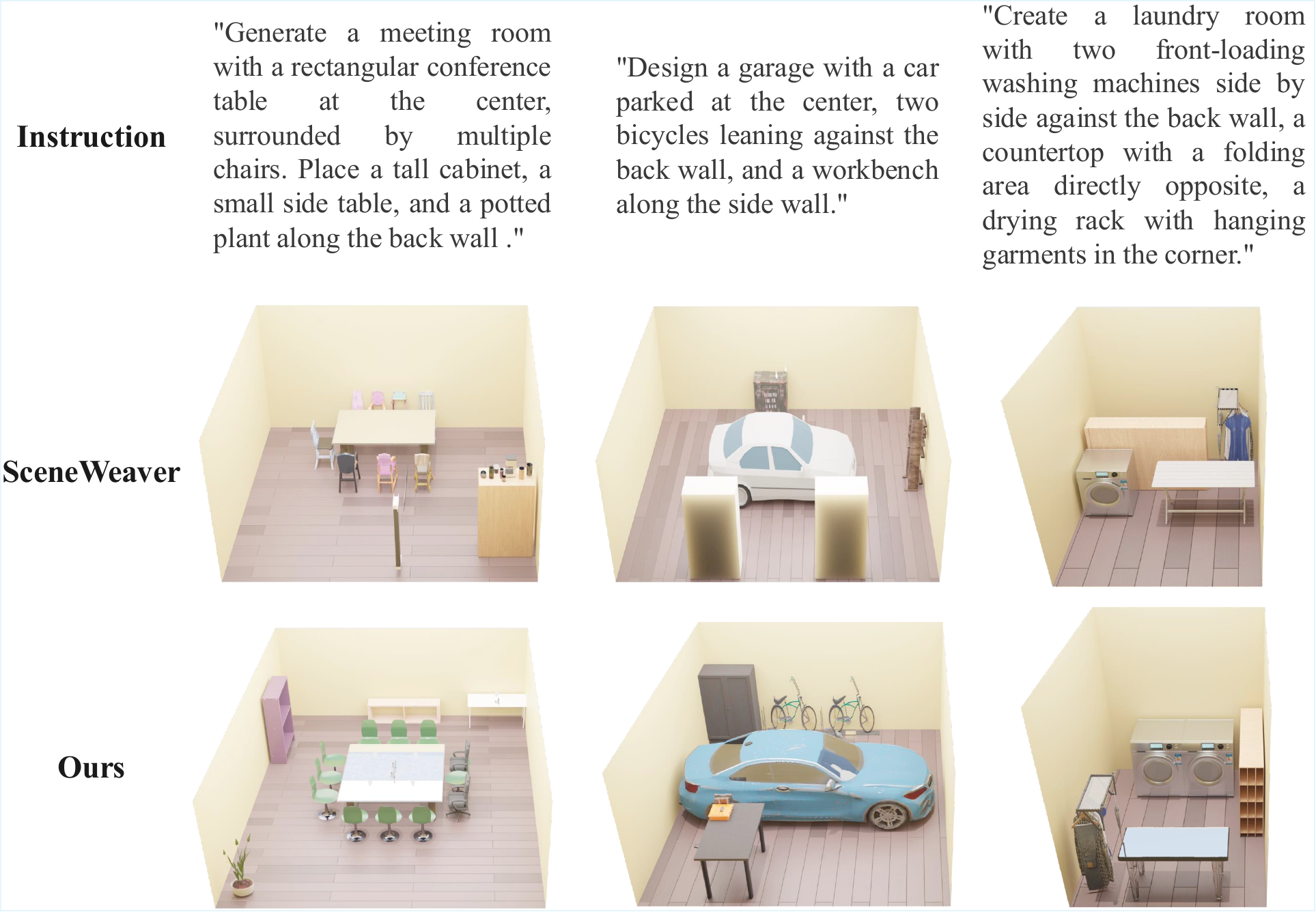}
    \caption{\textbf{Qualitative comparison on complex instructions.}  Our method follows instructions more faithfully and produces more realistic layouts. Please zoom in for better visualization.
    }
    \label{fig:complex}
    \vspace{-2.0em}
\end{figure*}

\vspace{-1em}
\mypara{Comparison on Complex Instructions.}
In addition to template-based instructions, we further evaluate our method and SceneWeaver on 70 complex instructions, each involving an unseen room type. 
These complex instructions specify 
object counts, positions, spatial relationships, and overall layout, providing a strong test of the model's fine-grained control capability during generation.
We report the quantitative results in Tab. \ref{tab:complex}, where each instruction is executed three times and the results are averaged over all instructions.
We observe that even on the more challenging complex instructions, our method consistently outperforms SceneWeaver on both physical and visual metrics. 
We also provide qualitative comparisons in Fig. \ref{fig:complex}, showing that our method follows complex instructions more faithfully and produces more plausible, realistic layouts.

\vspace{-1.5em}
\begin{table}[htbp]
  \centering
  \caption{\textbf{Quantitative comparison on complex instructions.} Our method consistently outperforms SceneWeaver, even on more challenging instructions.
}
  \vspace{-0.5em}
  \label{tab:complex}
  \setlength{\tabcolsep}{5pt}
  \renewcommand{\arraystretch}{1.05}
  \begin{adjustbox}{max width=0.84\linewidth}
  \begin{tabular}{l cccccccc}
    \toprule
    \textbf{Method} & \textbf{\#Obj $\uparrow$} & \textbf{\#OB $\downarrow$} & \textbf{\#CN $\downarrow$} & \textbf{Real. $\uparrow$} & \textbf{Func. $\uparrow$} & \textbf{Lay. $\uparrow$} & \textbf{Comp. $\uparrow$} & \textbf{Time (min) $\downarrow$} \\
    \midrule
    SceneWeaver \cite{yang2025sceneweaver} & 24.3 & 0.2 & 0.4 & 8.2 & 7.6 & 6.7 & 7.3 & 82.6 \\
    \textbf{Ours} & \textbf{27.6} & \textbf{0.1} & \textbf{0.2} & \textbf{8.6} & \textbf{8.3} & \textbf{7.2} & \textbf{7.7} & \textbf{46.9}  \\
    \bottomrule
  \end{tabular}
  \end{adjustbox}
  \vspace{-2.5em}
\end{table}

\vspace{-1em}
\subsection{Ablation Studies}
\vspace{-0.5em}
\mypara{Ablation Study on Training Strategy.}
To demonstrate the effectiveness of our two-phase training strategy, we evaluate the following variants:
\begin{itemize}
  \setlength{\itemsep}{0pt}
  \setlength{\parsep}{0pt}
  \setlength{\parskip}{0pt}
  \item \textit{Orchestrator without DPO training}: the orchestrator is trained only with stepwise SFT and trajectory-level SFT (first row in Tab.~\ref{tab:ablation_strategy}).
  \item \textit{Orchestrator without Stepwise Training}: the orchestrator is trained only with trajectory-level SFT and trajectory-level DPO (second row in Tab.~\ref{tab:ablation_strategy}).
  \item \textit{Orchestrator without Discriminator}: we train the orchestrator with stepwise SFT, trajectory-level SFT, stepwise DPO, and trajectory-level DPO, but remove the discriminator (third row in Tab.~\ref{tab:ablation_strategy}).
  \item \textit{Independent Training Only}: we perform the full independent training but skip interleaved training. At test time, the orchestrator samples multiple trajectories for each test instruction, and the discriminator selects the best one for execution (fourth row in Tab.~\ref{tab:ablation_strategy}).
\end{itemize}

As shown in Tab. \ref{tab:ablation_strategy}, DPO training, stepwise training, the discriminator, and interleaved training all contribute positively to the final results. Specifically, DPO training enhances the orchestrator's ability to predict full trajectories through preference learning; stepwise training equips the orchestrator with context-aware tool selection capabilities; the discriminator possesses explicit ranking ability to select the optimal trajectory from multiple candidates; and interleaved training allows the discriminator to adapt to the orchestrator's evolving trajectory distribution while distilling its discrimination capabilities back to the orchestrator.
Note that we do not evaluate an ``Orchestrator without trajectory-level training" variant (i.e., S-SFT + S-DPO) because, in this setting, the orchestrator can only learn next-step tool selection without acquiring the ability to predict complete trajectories, which is inconsistent with our inference process.

\vspace{-1.0em}
\begin{table}[htbp]
  \centering
  \caption{\textbf{Ablation study on training strategy.}
  S-SFT: stepwise SFT; T-SFT: trajectory-level SFT;
  S-DPO: stepwise DPO; T-DPO: trajectory-level DPO;
  Disc.: discriminator; Inter.: interleaved training. All results are based on generic template-based prompts. Each component contributes positively to the final results.}
  \vspace{-0.5em}
  \label{tab:ablation_strategy}
  \setlength{\tabcolsep}{4pt}
  \renewcommand{\arraystretch}{1.15}
  \begin{adjustbox}{max width=\linewidth}
  \begin{tabular}{l cccccc|ccccccc}
    \toprule
    \textbf{Setting} &
    \textbf{S-SFT} & \textbf{T-SFT} & \textbf{S-DPO} & \textbf{T-DPO} & \textbf{Disc.} & \textbf{Inter.} &
    \textbf{\#Obj $\uparrow$} & \textbf{\#OB $\downarrow$} & \textbf{\#CN $\downarrow$} & \textbf{Real. $\uparrow$} & \textbf{Func. $\uparrow$} & \textbf{Lay. $\uparrow$} & \textbf{Comp. $\uparrow$} \\
    \midrule
    w/o DPO           & \cmark & \cmark &       &       &       &       & 24.9 & 0.0 & 0.2 & 8.0 & 7.5 & 7.0 & 7.0 \\
    w/o Stepwise      &       & \cmark &       & \cmark &       &       & 26.2 & 0.0 & 0.1 & 8.2 & 8.0 & 7.3 & 7.2 \\
    w/o Discriminator & \cmark & \cmark & \cmark & \cmark &       &       & 27.6 & 0.0 & 0.0 & 8.4 & 8.4 & 7.5 & 7.4 \\
    Indep. only       & \cmark & \cmark & \cmark & \cmark & \cmark &       & 29.7 & 0.0 & 0.0 & 8.8 & 8.7 & 7.7 & 7.8 \\
    \textbf{Full}              & \cmark & \cmark & \cmark & \cmark & \cmark & \cmark & \textbf{31.1} & \textbf{0.0} & \textbf{0.0} & \textbf{9.1} & \textbf{9.1} & \textbf{7.9} & \textbf{8.1} \\
    \bottomrule
  \end{tabular}
  \end{adjustbox}
  \vspace{-2.0em}
\end{table}

\vspace{-0.5em}
\mypara{Ablation Study on Orchestrator.}
To further validate the necessity of orchestrator fine-tuning and demonstrate the efficiency of our training strategy, we compare against off-the-shelf LLM orchestrators. 
Specifically, we evaluate GPT-5.5 \cite{gpt5.5}, Gemini-3.1-Pro-Preview \cite{gemini3.1pro} and Qwen3-4B-Instruct-2507 \cite{yang2025qwen3} in an \textit{in-context learning} setting \cite{brown2020language}.
By providing these models with extensive instructions paired with high-quality, complete trajectories (sourced from our trajectory-level SFT training data), we enable the models to learn the mapping between instructions and high-quality trajectories. 
We then feed each test instruction and ask the model to predict the full tool-call trajectory.

\begin{table}[htbp]
  \centering
  \caption{\textbf{Ablation study on orchestrator.} We use off-the-shelf LLMs as orchestrators in an in-context learning setting, but they perform far worse than our fine-tuned orchestrator. All results are based on generic template-based prompts.
}
  \vspace{-0.5em}
  \label{tab:llm_avg}
  \setlength{\tabcolsep}{5pt}
  \renewcommand{\arraystretch}{1.05}
  \begin{adjustbox}{max width=0.84\linewidth}
  \begin{tabular}{l cccccccc}
    \toprule
    \textbf{Method} & \textbf{\#Obj $\uparrow$} & \textbf{\#OB $\downarrow$} & \textbf{\#CN $\downarrow$} & \textbf{Real. $\uparrow$} & \textbf{Func. $\uparrow$} & \textbf{Lay. $\uparrow$} & \textbf{Comp. $\uparrow$} & \textbf{Time (min)$\downarrow$} \\
    \midrule
    GPT-5.5 \cite{gpt5.5} & 24.2 & 0.0 & 0.3 & 8.1 & 7.7 & 7.1 & 7.4 & 45.3 \\
    Gemini-3.1 \cite{gemini3.1pro}  & 20.7 & 0.0 & 0.2 & 8.1 & 7.6 & 7.0 & 7.3 & 40.9 \\
    Qwen3 (w/o Finetune) \cite{yang2025qwen3} & 18.2 & 0.3 & 0.3 & 7.7 & 7.1 & 6.9 & 7.0 & 43.6 \\
    \textbf{Ours} & \textbf{31.1} & \textbf{0.0} & \textbf{0.0} & \textbf{9.1} & \textbf{9.1} & \textbf{7.9} & \textbf{8.1} & \textbf{39.7} \\
    \bottomrule
  \end{tabular}
  \end{adjustbox}
  \vspace{-1.5em}
\end{table}

The results in Tab. \ref{tab:llm_avg} show that, even when provided with a large number of example pairs in the context, these models still struggle to learn how to generate high-quality trajectories. 
In contrast, our fine-tuned models acquire strong trajectory generation and discrimination capabilities, leading to a substantial improvement in generation quality.

\vspace{-1em}

\subsection{User Study}
\vspace{-0.2em}
To further validate our evaluation metrics with human judgments, we conduct a user study on generated scene quality. 
We randomly collect 30 generated scenes from our method and the baselines, and ask 16 participants to rate them on a 1--10 scale across four dimensions: realism, functionality, layout, and completeness. 
The scoring criteria are the same as those used by the judge model (GPT-4.1 \cite{achiam2023gpt}). 
As shown in Tab.~\ref{tab:user_study_rating}, our method achieves higher human scores than all baselines across all dimensions. 
We also conduct an additional pairwise user study to evaluate human preferences in terms of generation quality and diversity.
Our user study received an IRB exemption from our university, and all participants provided informed consent.
Please refer to Fig.~\ref{fig:irb_exemption} and Fig.~\ref{fig:consent} for details.

\vspace{-1.5em}
\begin{table}[htbp]
  \centering
  \caption{\textbf{Human rating results.} Human participants rate generated scenes on a 1--10 scale across realism, functionality, layout, and completeness. Our method achieves the highest scores across all dimensions.}
  \vspace{-1em}
  \label{tab:user_study_rating}
  \setlength{\tabcolsep}{4pt}
  \renewcommand{\arraystretch}{1.05}
  \small
  \resizebox{0.75\columnwidth}{!}{%
  \begin{tabular}{lcccc}
    \toprule
    \textbf{Method} 
    & \textbf{Real. (Human) $\uparrow$} 
    & \textbf{Func. (Human) $\uparrow$} 
    & \textbf{Lay. (Human) $\uparrow$} 
    & \textbf{Comp. (Human) $\uparrow$} \\
    \midrule
    LayoutGPT~\cite{feng2023layoutgpt}
      & 3.01 & 2.97 & 2.84 & 3.25 \\
    Holodeck~\cite{yang2024holodeck}
      & 5.57 & 5.78 & 5.29 & 5.88 \\
    I-Design~\cite{ccelen2024design}
      & 4.73 & 4.98 & 4.33 & 4.64 \\
    SceneWeaver~\cite{yang2025sceneweaver}
      & 6.72 & 6.83 & 6.41 & 6.57 \\
    \textbf{Ours}
      & \textbf{7.60} & \textbf{7.99} & \textbf{7.46} & \textbf{7.76} \\
    \bottomrule
  \end{tabular}%
  }
  \vspace{-3.5em}
\end{table}

\section{Conclusion}
\vspace{-0.2em}
We propose a trainable orchestration framework for efficient agentic scene synthesis. 
It consists of an orchestrator and a discriminator. 
The orchestrator performs context-aware tool selection and predicts complete tool-call trajectories, while the discriminator ranks trajectories and selects the best one from multiple candidates. 
We further introduce a two-phase training strategy. In the first phase, we train the orchestrator and discriminator separately to align their output formats and build initial generation and discrimination abilities. 
Specifically, the orchestrator is trained in four stages: stepwise SFT, trajectory-level SFT, stepwise DPO, and trajectory-level DPO.
And the discriminator learns trajectory-level assessment via SFT.
In the second phase, we perform interleaved training: the discriminator first adapts to the orchestrator’s evolving trajectory distribution and then distills its discrimination capability back to the orchestrator, eliminating the need for the discriminator at inference time. 
Extensive experiments show that, compared to the state-of-the-art~\cite{yang2025sceneweaver}, our method produces higher-quality results while significantly reducing execution time.

\mypara{Acknowledgements.}
This research is based upon work supported by the Office of the Director of National Intelligence (ODNI), Intelligence Advanced Research Projects Activity (IARPA), via IARPA R\&D Contract No. 140D0423C0076. The views and conclusions contained herein are those of the authors and should not be interpreted as necessarily representing the official policies or endorsements, either expressed or implied, of the ODNI, IARPA, or the U.S. Government. The U.S. Government is authorized to reproduce and distribute reprints for Governmental purposes notwithstanding any copyright annotation thereon.

\clearpage  

%
%
\bibliographystyle{splncs04}
\bibliography{main}

\clearpage

In the supplementary material, we provide additional details and analyses, including a user study evaluating scene quality and diversity, IRB exemption and informed consent information for the user studies, an ablation study on SceneWeaver's orchestrator, a detailed runtime analysis, implementation details, additional qualitative results, and a discussion of limitations.

\vspace{-1em}
\section{Additional User Study}
\begin{figure*}[htbp]
    \centering
    \includegraphics[width=0.8\linewidth]{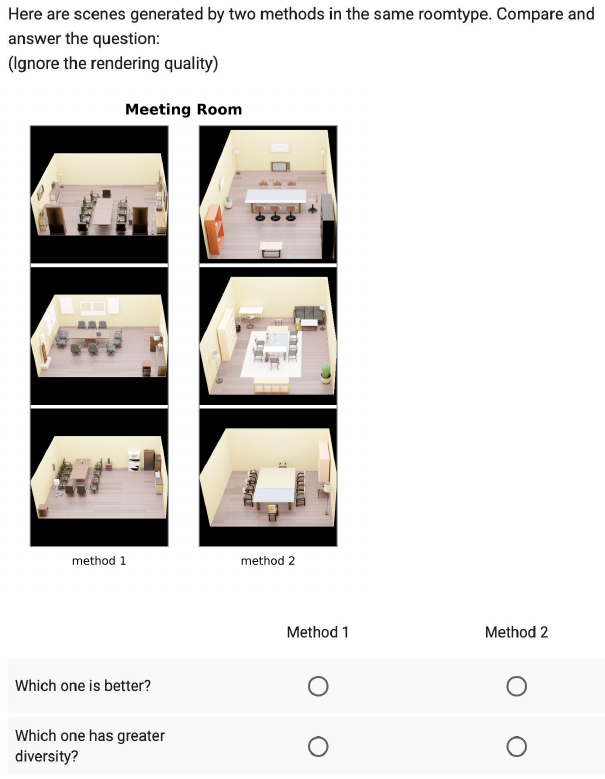}
    \caption{\textbf{Example of pairwise user study.}  
    }
    \label{fig:example_of_user_study}
    \vspace{-1.0em}
\end{figure*}

To further evaluate the \textit{quality} and \textit{diversity} of generated scenes, we conduct a user study with 18 participants. 
Following SceneWeaver \cite{yang2025sceneweaver}, we perform pairwise comparisons. 
Specifically, for each baseline, we randomly select 5 room types with 3 scenes per room type. 
In each trial, participants are presented with rendered images of 3 scenes generated by our method and 3 scenes generated by the baseline, where both methods are conditioned on the same room types, as shown in Fig. \ref{fig:example_of_user_study}. 
Participants are then asked to answer two questions: which method produces higher-quality results that they prefer, and which method produces more diverse results. 
We collect all responses and compute the average across participants.
As shown in Tab. \ref{tab:user_study}, our method generates higher-quality and more diverse scenes than the baselines in the vast majority of cases.

\vspace{-1.0em}
\begin{table}[htbp]
  \centering
  \caption{\textbf{User study results.} Our method generates higher-quality and more diverse scenes than the baselines.}
  \vspace{-0.5em}
  \label{tab:user_study}
  \setlength{\tabcolsep}{5pt}
  \renewcommand{\arraystretch}{1.05}
  \begin{adjustbox}{max width=0.98\linewidth}
  \begin{tabular}{l cccc}
    \toprule
    \textbf{Metric} & \textbf{w.r.t LayoutGPT} \cite{feng2023layoutgpt} & \textbf{w.r.t Holodeck} \cite{yang2024holodeck} & \textbf{w.r.t I-Design} \cite{ccelen2024design} & \textbf{w.r.t SceneWeaver} \cite{yang2025sceneweaver} \\
    \midrule
    Preference & 100\% & 84.4\% & 95.6\% & 76.7\% \\
    Diversity  & 100\% & 88.9\% & 96.6\% & 80.8\% \\
    \bottomrule
  \end{tabular}
  \end{adjustbox}
  \vspace{-1.5em}
\end{table}

\begin{figure*}[htbp]
    \centering
    \includegraphics[width=0.85\linewidth]{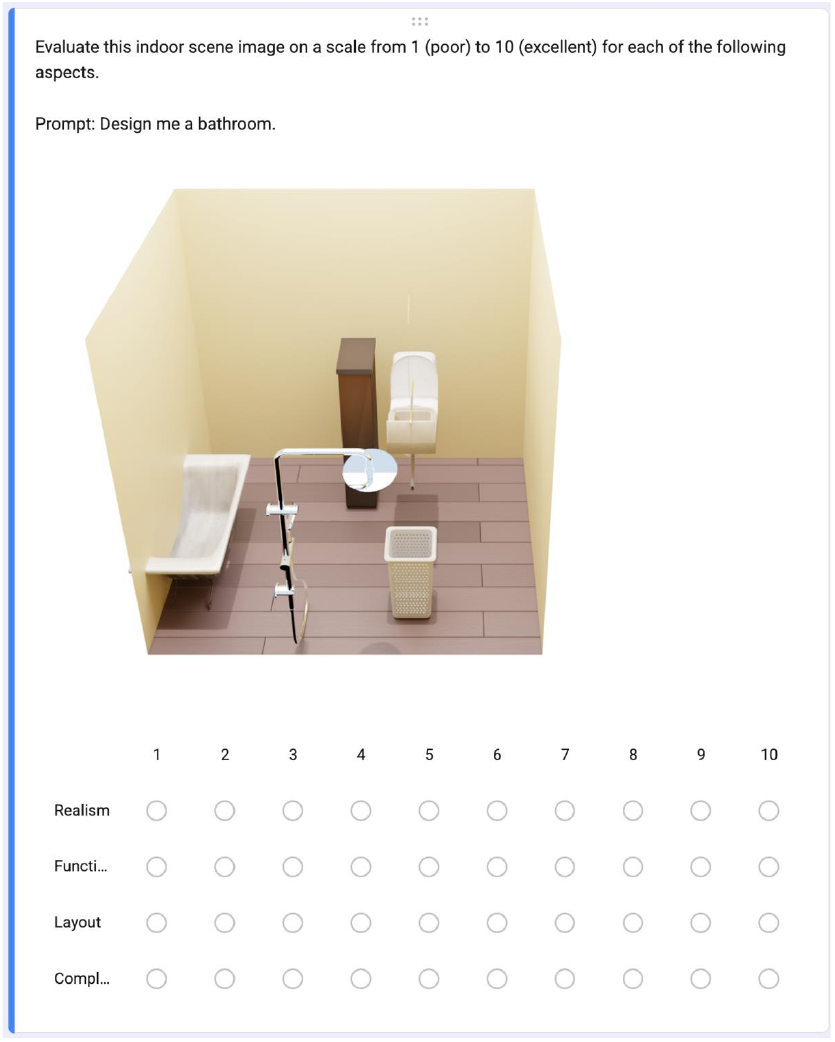}
    \caption{\textbf{Example of human rating user study.}  
    }
    \label{fig:example_of_human_rating_user_study}
    \vspace{-1.0em}
\end{figure*}

For the human rating user study (Tab.~\ref{tab:user_study_rating}), we also provide an example in Fig.~\ref{fig:example_of_human_rating_user_study}. Based on the human ratings and GPT-4.1 \cite{achiam2023gpt} judge scores, we further measure their alignment by converting both sets of scores into rankings over all sampled scenes and computing Kendall's~$\tau$. 
As shown in Tab.~\ref{tab:user_gpt_agreement}, the results indicate strong user-user and user-GPT-4.1 agreement across all metrics.

\vspace{-0.5em}
\begin{table}[htbp]
  \centering
  \caption{\textbf{Agreement between human judgments and GPT-4.1.} We compute Kendall's~$\tau$ over scene-level rankings to measure both user-user agreement and user-GPT-4.1 alignment across all sampled scenes.}
  \vspace{-0.5em}
  \label{tab:user_gpt_agreement}
  \setlength{\tabcolsep}{5pt}
  \renewcommand{\arraystretch}{1.05}
  \small
  \resizebox{0.55\columnwidth}{!}{%
  \begin{tabular}{lcccc}
    \toprule
    \textbf{Agreement}
    & \textbf{Real. $\uparrow$}
    & \textbf{Func. $\uparrow$}
    & \textbf{Lay. $\uparrow$}
    & \textbf{Comp. $\uparrow$} \\
    \midrule
    User--User
    & 0.49 & \textbf{0.53} & \textbf{0.53} & 0.50 \\
    User--GPT-4.1
    & \textbf{0.50} & 0.47 & 0.51 & \textbf{0.52} \\
    \bottomrule
  \end{tabular}}
  \vspace{-1.0em}
\end{table}

\section{IRB Exemption and Informed Consent}
\begin{figure*}[htbp]
    \centering
    \includegraphics[width=\linewidth]{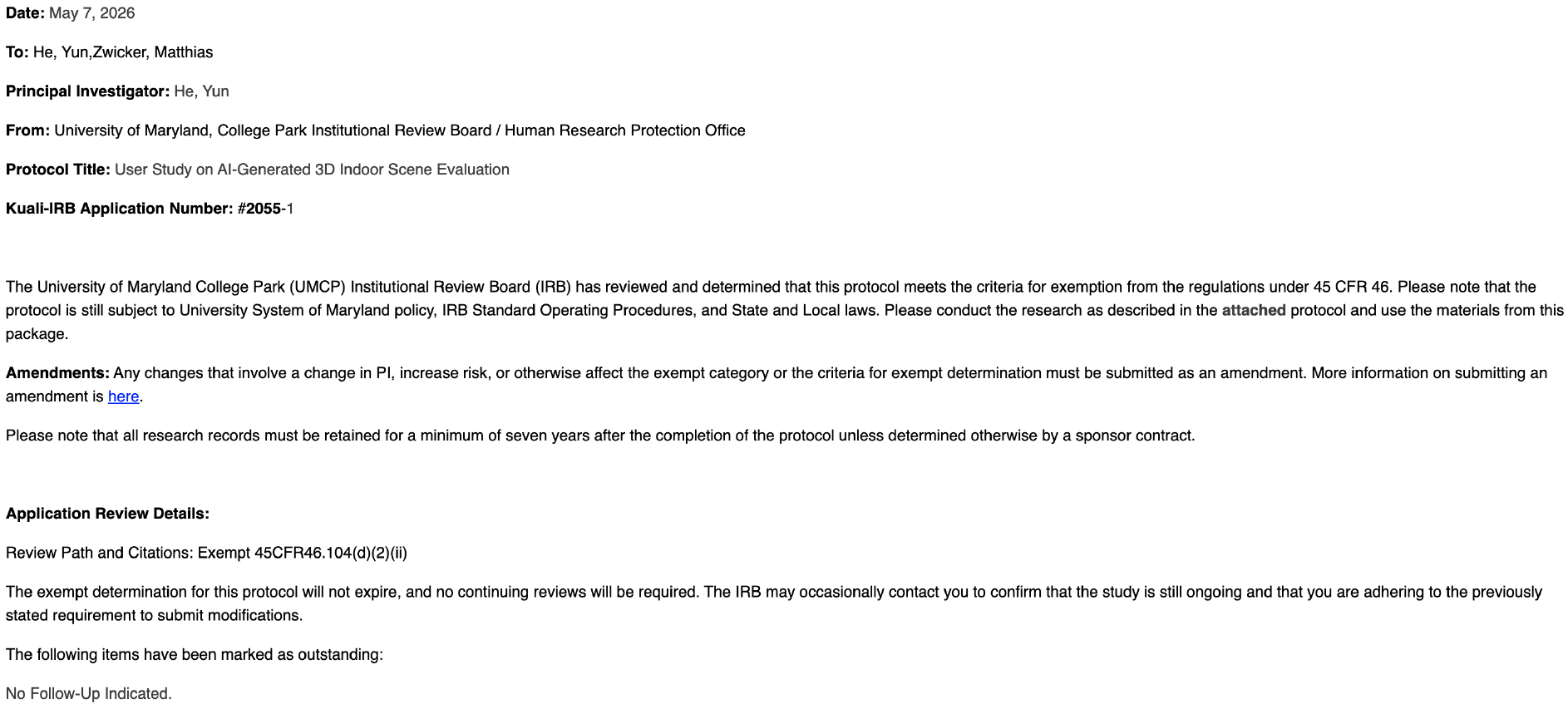}
    \caption{\textbf{IRB exemption.}  
    }
    \label{fig:irb_exemption}
    \vspace{-1.0em}
\end{figure*}

\begin{figure*}[htbp]
    \centering
    \includegraphics[width=0.75\linewidth]{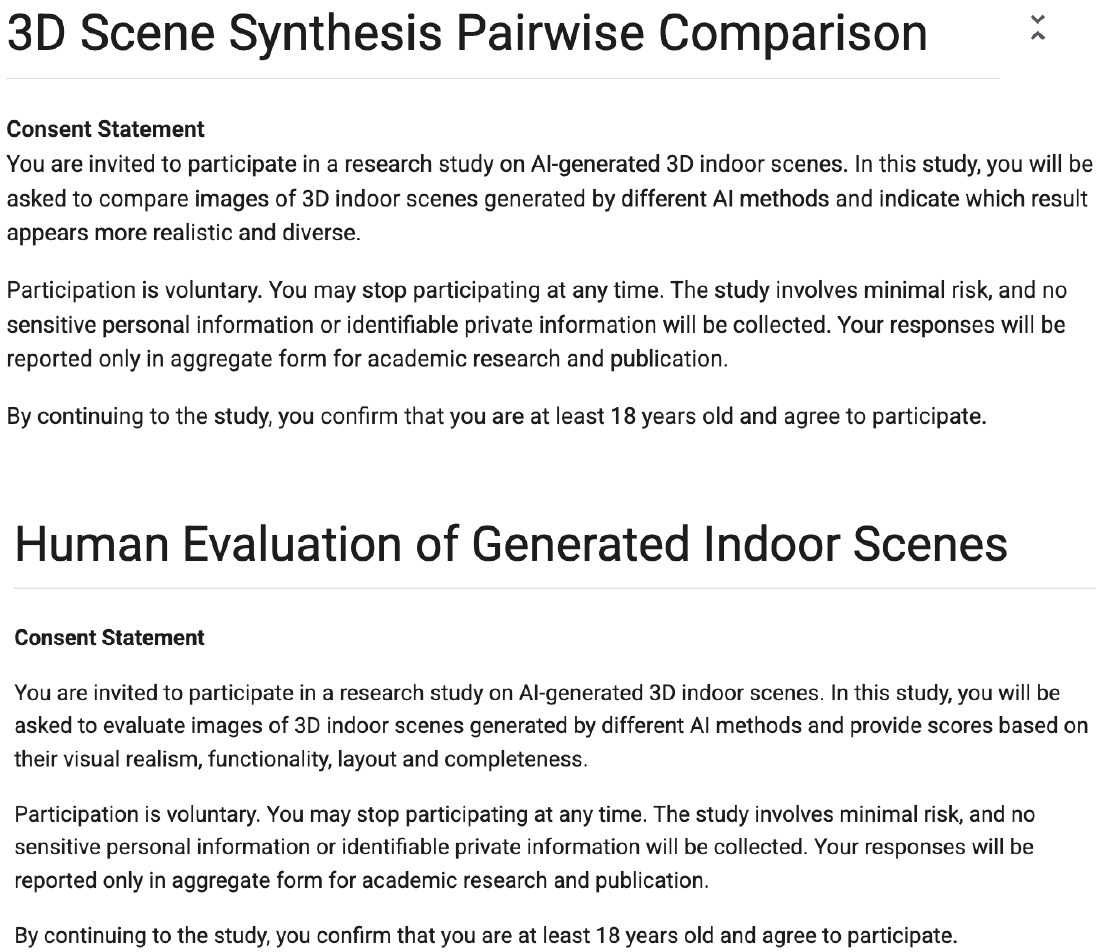}
    \caption{\textbf{Informed consent statements.}  
    }
    \label{fig:consent}
    \vspace{-1.0em}
\end{figure*}

Our user studies were reviewed by the Institutional Review Board (IRB) at the University of Maryland, College Park and determined to be exempt from IRB regulations under Kuali-IRB application \#2055-1, as shown in Fig.~\ref{fig:irb_exemption}. 
Before participating, all users were presented with a consent statement explaining the purpose of the study, voluntary participation, minimal risk, and aggregate-only reporting of responses, as shown in Fig.~\ref{fig:consent}.

\section{Ablation on SceneWeaver's Orchestrator}
To further validate the effectiveness of our tool orchestration framework, we directly replace SceneWeaver's \cite{yang2025sceneweaver} original orchestrator with two stronger off-the-shelf LLM models, GPT-5.5 \cite{gpt5.5} and Gemini-3.1-Pro-Preview \cite{gemini3.1pro}, for step-by-step tool selection, while keeping all other components unchanged. 
As shown in Tab.~\ref{tab:strong_llm}, even when equipped with these stronger off-the-shelf models as the orchestrator, the resulting performance remains clearly behind our fine-tuned orchestrator across all metrics, indicating that the gains come from our trainable orchestration framework.

\vspace{-0.5em}
\begin{table}[htbp]
  \centering
  \caption{\textbf{Ablation on SceneWeaver's orchestrator.} We replace SceneWeaver's original orchestrator with two stronger off-the-shelf models, GPT-5.5 and Gemini-3.1-Pro-Preview, while keeping all other components unchanged. Even with these stronger models as the orchestrator, the performance remains clearly behind our fine-tuned orchestrator across all metrics. All results are based on generic template-based prompts.}
  \vspace{-0.5em}
  \label{tab:strong_llm}
  \setlength{\tabcolsep}{2.0pt}
  \renewcommand{\arraystretch}{0.95}
  \scriptsize
  \resizebox{0.85\columnwidth}{!}{%
  \begin{tabular}{l cccccccc}
    \toprule
    \textbf{Method}
    & \textbf{\#Obj$\uparrow$}
    & \textbf{\#OB$\downarrow$}
    & \textbf{\#CN$\downarrow$}
    & \textbf{Real.$\uparrow$}
    & \textbf{Func.$\uparrow$}
    & \textbf{Lay.$\uparrow$}
    & \textbf{Comp.$\uparrow$}
    & \textbf{Time (min)$\downarrow$} \\
    \midrule
    Gemini-3.1 \cite{gemini3.1pro}
      & 28.5 & 0.1 & 0.5 & 8.9 & 8.8 & 7.3 & 7.6 & 103.5 \\
    GPT-5.5 \cite{gpt5.5}
      & 26.4 & 0.0 & 0.1 & 8.8 & 8.9 & 7.2 & 7.5 & 82.6 \\
    \textbf{Ours}
      & \textbf{31.1} & \textbf{0.0} & \textbf{0.0}
      & \textbf{9.1} & \textbf{9.1} & \textbf{7.9} & \textbf{8.1} & \textbf{39.7} \\
    \bottomrule
  \end{tabular}}
  \vspace{-3.0em}
\end{table}

\begin{figure*}[t!]
    \centering
    \includegraphics[width=\linewidth]{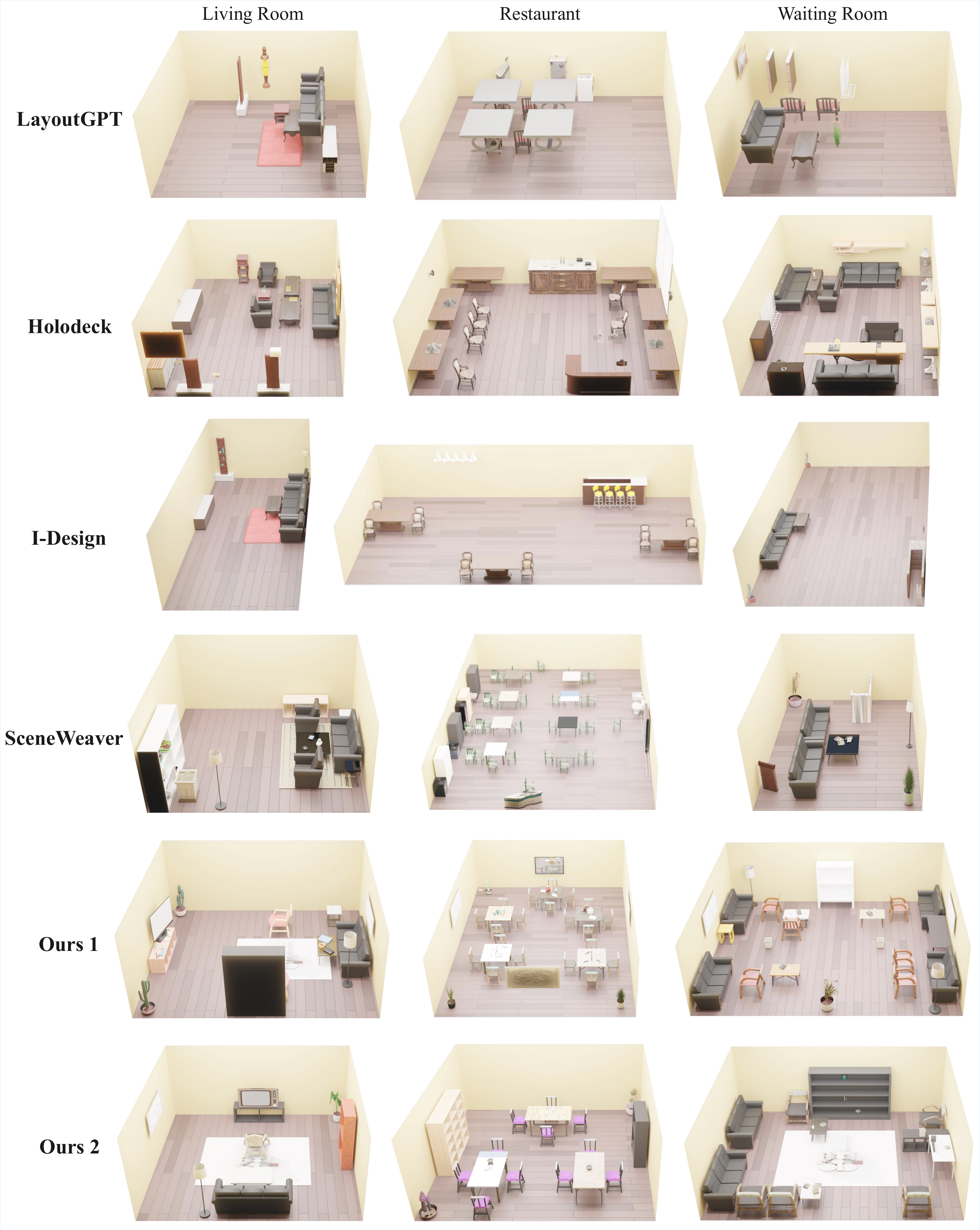}
    \caption{\textbf{Extra qualitative comparison with baselines} on both seen (waiting
    room) and unseen (living room, restaurant) room types. 
    ``Ours 1'' and ``Ours 2'' denote two different scenes generated by our method from the same room type. Our method not only produces more realistic scenes, but also generates diverse results.
    }
    \label{fig:base}
    \vspace{-1.0em}
\end{figure*}

\begin{figure*}[t!]
    \centering
    \includegraphics[width=\linewidth]{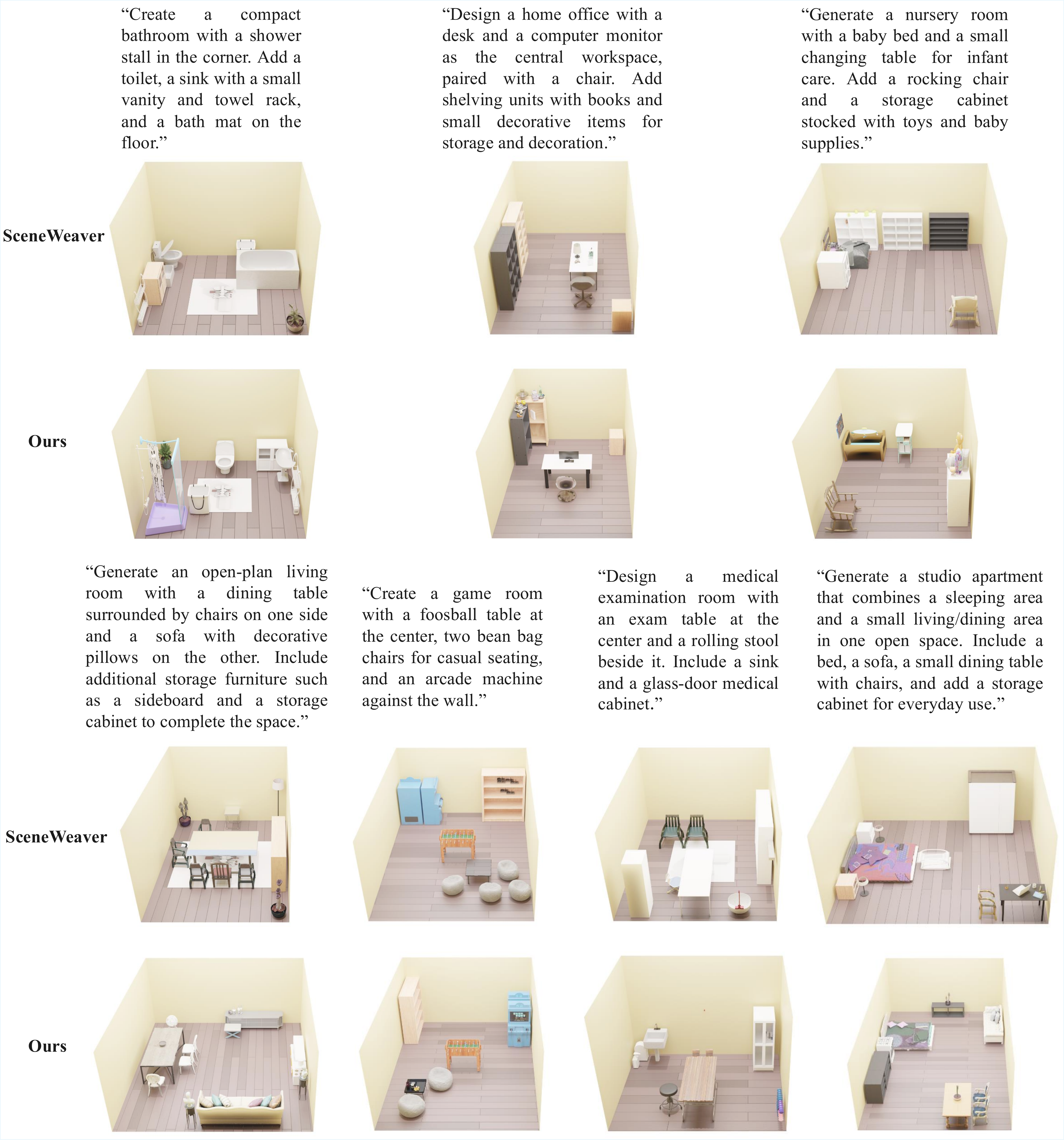}
    \caption{\textbf{Extra qualitative results on complex instructions.}  Our method follows instructions more faithfully and produces more realistic layouts. 
    }
    \label{fig:supp_complex}
    \vspace{-0.5em}
\end{figure*}

\section{Detailed Runtime Analysis}
In Tab.~\ref{tab:generation_time}, we report the average generation time of each method across all room types. 
Here, we further provide the detailed generation time for each method on each room type. 
As shown in Tab.~\ref{tab:generation_time_per_room}, our method substantially reduces the generation time of agentic methods by optimizing the execution flow and removing the step-by-step review process.

\vspace{-0.5em}
\begin{table*}[htbp]
\centering
\caption{\textbf{Per-room generation time comparison (minutes).} We report the generation time for each room type. All methods run on a single A6000 GPU. Monolithic methods are much faster than agentic approaches, since agentic methods always invoke multiple tools. By optimizing the execution flow and removing the step-by-step review process, our method substantially reduces the runtime of agentic generation.}
\vspace{-0.5em}
\label{tab:generation_time_per_room}
\setlength{\tabcolsep}{5pt}
\renewcommand{\arraystretch}{1.05}
\begin{adjustbox}{max width=0.7\linewidth}
\begin{tabular}{lccccc}
\toprule
\textbf{Method} &
\textbf{Bedroom} &
\textbf{Living Room} &
\textbf{Kitchen} &
\textbf{Gym} &
\textbf{Restaurant} \\
\midrule
LayoutGPT \cite{feng2023layoutgpt}      & 2.5 & 2.3 & 2.9 & 2.9 & 1.9 \\
Holodeck \cite{yang2024holodeck}        & 5.3 & 17.7 & 7.4 & 22.5 & 10.1 \\
I-Design \cite{ccelen2024design}        & 2.6 & 2.7 & 8.2 & 7.5 & 14.9 \\
SceneWeaver \cite{yang2025sceneweaver}  & 42.0 & 62.8 & 122.8 & 40.4 & 289.4 \\
Ours                                    & 19.6 & 31.2 & 81.3 & 19.6 & 81.2 \\
\bottomrule
\end{tabular}
\end{adjustbox}
\vspace{0.35em}
\begin{adjustbox}{max width=0.85\linewidth}
\begin{tabular}{lccccc}
\toprule
\textbf{Method} &
\textbf{Bathroom} &
\textbf{Children Room} &
\textbf{Meeting Room} &
\textbf{Office} &
\textbf{Waiting Room} \\
\midrule
LayoutGPT \cite{feng2023layoutgpt}      & 1.9 & 1.9 & 2.6 & 2.5 & 2.4 \\
Holodeck \cite{yang2024holodeck}        & 3.5 & 8.1 & 14.9 & 17.3 & 17.6 \\
I-Design \cite{ccelen2024design}        & 2.5 & 8.3 & 4.9 & 4.3 & 4.8 \\
SceneWeaver \cite{yang2025sceneweaver}  & 58.5 & 97.9 & 81.0 & 52.5 & 62.4 \\
Ours                                    & 22.4 & 36.0 & 52.3 & 23.6 & 29.5 \\
\bottomrule
\end{tabular}
\end{adjustbox}
\label{tab:time_for_each_roomtype}
\vspace{-1.0em}
\end{table*}

\section{Implementation Details}
For composition score calculation in Eq. \ref{eq:quality} and Eq. \ref{eq:composition} of the main paper, we set $\alpha = 4$, 
$\lambda = 0.1$, and $\gamma = 0.05$. During independent training, we use the 
following thresholds: 1) Stepwise SFT: $\tau_1 = 3$; 2) Trajectory-level SFT: $\tau_2 = 7.5$; 3) Stepwise DPO: $\tau_3 = 3$; 4) Trajectory-level DPO: $\tau_4 = 3$.
For interleaved training, we run only a single cycle, i.e., we train the discriminator and the orchestrator only once.

\section{More Qualitative Comparison}
In Fig. \ref{fig:base}, we provide additional qualitative comparisons between our method and the baselines. 
We also present multiple scenes generated by our method within the same room type. 
Overall, our method consistently outperforms the baselines across all room types, producing more physically plausible layouts with better spatial organization and higher visual realism, while also generating diverse scenes. 
In contrast, LayoutGPT \cite{feng2023layoutgpt} often yields poor physical plausibility, with furniture floating in midair. 
Holodeck \cite{yang2024holodeck} tends to generate many objects, but the placements are cluttered and the layouts are unrealistic. 
I-Design \cite{ccelen2024design} produces scenes with fewer objects, resulting in lower completeness.
SceneWeaver \cite{yang2025sceneweaver} can generate generally realistic scenes, but it sometimes misses important details—for example, a living room without a TV, or a restaurant that contains a toilet and randomly placed chairs.

\section{More Results on Complex Instructions}
In Tab. \ref{tab:complex} and Fig. \ref{fig:complex}, we compare our method against SceneWeaver \cite{yang2025sceneweaver} on complex instructions. 
To obtain the quantitative results in Tab. \ref{tab:complex} of the main paper, we run each instruction 3 times and average the results, then compute the overall average across all instructions. 
Additionally, Fig. \ref{fig:supp_complex} provides extra qualitative results, showing that our method more accurately follows user instructions, produces more reasonable layouts, and generates more realistic scenes.

\section{Limitation}
Our method has three main limitations. 
First, the orchestrator can occasionally hallucinate when generating complete tool-call trajectories, leading to formatting errors. For example, it may output an incorrect tool name (e.g., ``addCrowd'' instead of the actual ``add\_crowd''), which prevents the system from matching and invoking the intended tool. A simple fix is to validate the generated trajectory and, if any formatting error is detected, regenerate it.
Second, our method is trained on trajectories generated within the SceneWeaver \cite{yang2025sceneweaver} framework. 
While our approach optimizes tool orchestration under this framework, its performance is ultimately bounded by the underlying tools. 
In other words, if the tools in SceneWeaver are not strong enough, our overall performance will also be limited. 
For example, since SceneWeaver does not support articulated objects in the generated scenes, our method cannot produce them either.
That said, our approach is designed as a general tool orchestration framework and should be applicable to other agentic workflows; if those workflows integrate stronger tools, our orchestrator could achieve better performance.
Third, our system relies on an underlying layout solver that performs constrained optimization (inherited from SceneWeaver). In crowded scenes, this solver may fail to converge, leading to invalid or implausible layouts. Improving the robustness of the layout solver or incorporating stronger constraint-handling mechanisms could further improve our method in such challenging cases.

\vspace{2em}

\end{document}